\documentclass{article}
\usepackage{spconf,amsmath,graphicx}
\usepackage{amsfonts,bm}

\usepackage{caption}
\usepackage{subcaption}

\usepackage{booktabs}       
\usepackage{multirow,makecell}

\usepackage{comment}

\usepackage{xcolor}

\title{w2v-BERT: Combining Contrastive Learning and Masked Language Modeling for Self-Supervised Speech Pre-Training}
\name{Yu-An Chung$^{1,2}$\sthanks{Work done during an internship at Google Brain.}, Yu Zhang$^{2}$, Wei Han$^{2}$, Chung-Cheng Chiu$^{2}$, James Qin$^{2}$, Ruoming Pang$^{2}$, Yonghui Wu$^{2}$}
\address{$^{1}$MIT Computer Science and Artificial Intelligence Laboratory\\$^{2}$Google Brain\\
\small \texttt{\{andyyuan, ngyuzh, weihan, chungchengc, jamesqin, rpang, yonghui\}@google.com}}
\begin{document}
\ninept
\maketitle
\begin{abstract}
Motivated by the success of masked language modeling~(MLM) in pre-training natural language processing models, we propose w2v-BERT that explores MLM for self-supervised speech representation learning.
w2v-BERT is a framework that combines contrastive learning and MLM, where the former trains the model to discretize input continuous speech signals into a finite set of discriminative speech tokens, and the latter trains the model to learn contextualized speech representations via solving a masked prediction task consuming the discretized tokens.
In contrast to existing MLM-based speech pre-training frameworks such as HuBERT, which relies on an iterative re-clustering and re-training process, or vq-wav2vec, which concatenates two separately trained modules, w2v-BERT can be optimized in an end-to-end fashion by solving the two self-supervised tasks~(the contrastive task and MLM) simultaneously.
Our experiments show that w2v-BERT achieves competitive results compared to current state-of-the-art pre-trained models on the LibriSpeech benchmarks when using the Libri-Light~60k corpus as the unsupervised data.
In particular, when compared to published models such as conformer-based wav2vec~2.0 and HuBERT, our model shows~5\% to~10\% relative WER reduction on the test-clean and test-other subsets.
When applied to the Google's Voice Search traffic dataset, w2v-BERT outperforms our internal conformer-based wav2vec~2.0 by more than~30\% relatively.
\end{abstract}

\section{Introduction}
\label{sec:intro}
How to leverage large-scale unannotated speech to improve supervised automatic speech recognition~(ASR) performance has been a longstanding research problem.
To date, there have been two major streams for utilizing unlabeled speech data for tackling such a semi-supervised ASR task.

The first line of work is self-training~\cite{riloff2003learning,yarowsky1995unsupervised,scudder1965probability}, also known as pseudo-labeling, where the system starts with training a teacher model using initially available labeled data.
Next, the teacher model is used to label the unlabeled data.
The combined labeled and pseudo-labeled data are then used to train a student model.
The pseudo-labeling process can be repeated multiple times to improve the quality of the teacher model.
Self-training has been a practically useful and extensively studied technique in ASR~\cite{kahn2020self,synnaeve2020end,li2019semi,parthasarathi2019lessons,novotney2009analysis,zavaliagkos1998utilizing}.

The second direction of taking advantage of unlabeled speech data is unsupervised pre-training, or self-supervised pre-training.
In unsupervised pre-training, a model is first trained to complete a proxy task that is designed to consume only unlabeled data~(hence unsupervised).
Such proxy task is commonly believed to be capable of initializing the parameters of the model at a good starting point before it is being trained on the supervised data.
Significant recent research effort has been made to develop proxy tasks that allow models to perform well when the models are fine-tuned on ASR tasks~\cite{oord2018representation,chung2019unsupervised,schneider2019wav2vec,ling2020deep,chung2020generative,liu2020mockingjay,chung2020improved,liu2021tera,wang2020unsupervised,ling2020decoar,bai2021representation}.
There have also been studies that show that the gains brought by self-training and unsupervised pre-training are additive in downstream ASR~\cite{zhang2020pushing,xu2021self}.

\begin{figure}[htbp]
  \centering
  \includegraphics[width=0.5\textwidth]{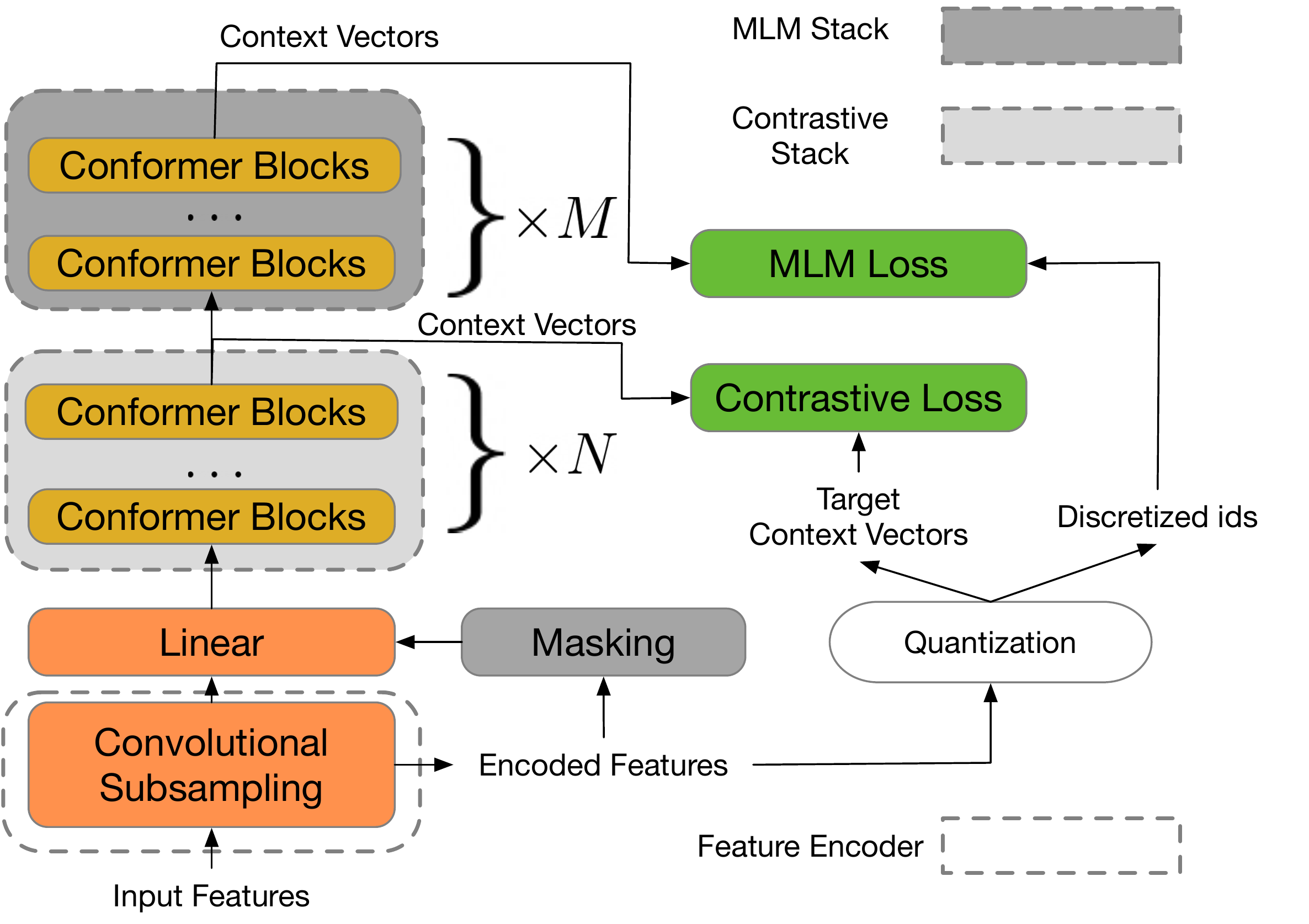}
  \caption{Illustration of the w2v-BERT pre-training framework. w2v-BERT is composed of a feature encoder, a contrastive module, and a masked language modeling~(MLM) module, where the latter two are both a stack of conformer blocks.~$N$ and~$M$ denote the number of conformer blocks in the two modules, respectively.}
  \label{fig:model}
\end{figure}

In this work, we focus on improving the unsupervised pre-training aspect of semi-supervised ASR by proposing a novel pre-training framework.
Our method, which we call w2v-BERT, combines the core methodologies from two recent frameworks for self-supervised pre-training of speech and language respectively: wav2vec~2.0~\cite{baevski2020wav2vec} and BERT~\cite{devlin2019bert}.
The idea of w2v-BERT is to use the contrastive task defined in wav2vec~2.0 to obtain an inventory of a finite set of discriminative, discretized speech units, and then use them as target in a masked prediction task in a way that is similar to masked language modeling~(MLM) proposed in BERT for learning contextualized speech representations.
Although the masked prediction task requires to consume tokens that are to be learned by solving the contrastive task first, we show that in practice the two objectives can be optimized simultaneously.
Figure~\ref{fig:model} illustrates the w2v-BERT pre-training framework.

In this paper, we make the following contributions:
\begin{itemize}
  \item We propose w2v-BERT that directly optimizes a contrastive loss and a masked prediction loss simultaneously for end-to-end self-supervised speech representation learning.
  \item We show that w2v-BERT yields state-of-the-art performance on the well-benchmarked LibriSpeech task.
  \item We show that w2v-BERT greatly improves a real-world recognition task~(voice search) over conformer-based wav2vec~2.0.
  \item We provide an analysis that empirically confirms the necessity of contrastive learning for enabling masked prediction in our framework. We also show in our voice search experiments that mask prediction is very useful for alleviating the problem of ``easy negative samples'' in contrastive learning.
\end{itemize}

The rest of the paper is organized as follows.
We begin with discussing the differences between w2v-BERT and some of the most relevant unsupervised speech pre-training frameworks from the literature in Section~\ref{sec:related}.
Then, in Section~\ref{sec:method} we present the w2v-BERT pre-training framework, including the model architecture and training objectives.
Section~\ref{sec:setup} describes the experimental setup, followed by our results and analysis in Section~\ref{sec:results} where we apply pre-trained w2v-BERT models to LibriSpeech and voice search ASR.
Finally, we conclude in Section~\ref{sec:conclusion}.

\section{Related Work}
\label{sec:related}
We consider our work most related to~HuBERT~\cite{hsu2021hubert}, vq-wav2vec~\cite{baevski2020vq}, and DiscreteBERT~\cite{baevski2019effectiveness}: w2v-BERT and these methods all try to first transform continuous speech signals into discretized units so as to exploit masked language modeling~(MLM)~\cite{devlin2019bert} for learning contextualized speech representations.
Despite sharing this same high-level philosophy for learning speech representations, there are two key differences between w2v-BERT and other methods.

The most noticeable difference is that w2v-BERT's speech discretizing module and its main contextualized representation learning module can be trained end-to-end.
This is in contrast to vq-wav2vec and DiscreteBERT, which involve a two-stage process where the speech discretizing module needs to be obtained in advance and is kept frozen during the training of the representation learning module.
In vq-wav2vec and DiscreteBERT, a problematic token ID assignment would negatively affect the subsequent learning module and it is hard for the learning module to recover the errors made by the discretizer.
Observing such drawback, HuBERT greatly improves vq-wav2vec and DiscreteBERT by allowing refinement on the ID assignment via iterating between k-means clustering and re-training its representation learning module. 
However, the fact that HuBERT iterates between the two stages also means it involves more heuristic design choices, for example, the gradually increasing number of clusters in different iterations.
End-to-end methods such as w2v-BERT alleviate the need of coordinating multiple stages.
One potential risk for end-to-end approaches compared to k-means clustering is codebook collapse.
In w2v-BERT, we find the contrastive learning objective effectively avoids codebook collapse and thus enables masked prediction training.

In addition, unlike other methods that use transformer layers~\cite{vaswani2017attention} as building blocks, w2v-BERT adopts conformer layers~\cite{gulati2020conformer} for constructing the network.
As demonstrated in~\cite{gulati2020conformer}, conformer layers, which combine convolution neural networks~(CNNs) and transformers to model both local and global dependencies of audio sequences, are likely a better option for modeling speech than transformer layers and CNNs.
That being said, using a potentially more powerful building block is not the only factor that makes w2v-BERT outperform other methods, as the effectiveness of the pre-training framework itself is also validated in our experiments where w2v-BERT outperforms w2v-Conformer~\cite{zhang2020pushing}, which is also built with conformer layers.

w2v-BERT is also related to wav2vec~2.0~\cite{baevski2020wav2vec}.
Same as w2v-BERT, wav2vec~2.0 is end-to-end where the discretizer is jointly trained with its representation learning module.
However, wav2vec~2.0 only employs contrastive learning, whose resulting ASR performance lags behind that of combining contrastive learning and masked prediction.

\section{Method}
\label{sec:method}
In this section we present each component in w2v-BERT, starting with its model architecture.

\subsection{Model architecture}

\begin{table*}[htbp]
  \caption{Parameters for w2v-BERT models. Dim. stands for dimension.}
  \vskip 0.1in
  \label{tab:model_params}
  \centering
  \resizebox{0.95\width}{!}{%
  \begin{tabular}{lcccccccccc}
    \toprule
    \bfseries Model & \# Params~(B) & \makecell{\# Contrastive\\Layers} & \makecell{\# Masked\\Layers} & \makecell{Model\\Dim.} & \makecell{Attention\\Heads} & \makecell{Conv. Layer\\Kernel Size} & \makecell{Relative\\Attention} & \makecell{Codebook\\Size} & \makecell{Code\\Dim.} \\
    \midrule
    w2v-BERT~XL     & 0.6 & 12 & 12 & 1024 & 8 & 5 & N & 1024 & 1024\\
    w2v-BERT~XXL    & 1.0 & 12 & 30 & 1024 & 8 & 5 & N & 1024 & 1024\\
    \bottomrule
  \end{tabular}
  }
\end{table*}
Our model architecture for pre-training is composed of a feature encoder that extracts latent speech representations from raw acoustic inputs, a module for solving wav2vec~2.0's contrastive task~\cite{baevski2020wav2vec} to obtain a set of discretized speech tokens, and a module for solving a masked prediction task~\cite{devlin2019bert} for learning contextualized speech representations.

\newcommand{\myparagraph}[1]{\vspace{.4em} \noindent \textbf{#1}\ }
\myparagraph{Feature encoder}
The feature encoder acts as a convolutional subsampling block that consists of two 2D-convolution layers, both with strides~$(2, 2)$, resulting in a 4x reduction in the acoustic input's sequence length.
Given, for example, a log-mel spectrogram as input, the feature encoder extracts latent speech representations that will be taken as input by the subsequent contrastive module.

\myparagraph{Contrastive module}
The module contains a linear projection layer followed by a stack of conformer blocks~\cite{gulati2020conformer}, each of which is a series of multi-headed self attention~\cite{vaswani2017attention}, depth-wise convolution and feed-forward layers.

The goal of the contrastive module is to discretize the feature encoder output into a finite set of representative speech units.
For this purpose, the contrastive module involves a quantization mechanism.
The output of the feature encoder, on one hand, is fed into the linear projection layer followed by the stack of conformer blocks {\it after masking} to produce context vectors, and on the other hand, is passed to the quantizer {\it without masking} to yield quantized vectors and their assigned token IDs.
The quantized vectors are used in conjunction with the context vectors that correspond to the masked positions to solve the contrastive task defined in wav2vec~2.0 to optimize the contrastive module; the assigned token IDs will be later used by the subsequent masked prediction module as prediction target.

\myparagraph{Masked prediction module}
The masked prediction module is a stack of conformer blocks where each block has an identical configuration to those from the contrastive module.
The module directly takes in the context vectors produced by the contrastive module and extracts high-level contextualized speech representations.

\subsection{Pre-training}
During pre-training only unlabeled speech data is used.

\myparagraph{Contrastive loss}
The contrastive loss is used to train the contrastive module along with the quantizer, such that the former yields adequate context vectors that will be taken as input by the subsequent masked prediction module, and the latter produces discriminative discretized speech tokens that will be used by the masked prediction module as prediction target.
We adopt the contrastive task defined in wav2vec~2.0 and follow its quantization mechanism.

Once the feature encoder has transformed the raw acoustic input into latent speech representations, we randomly select some time steps to mask.
Unlike wav2vec~2.0 where the masked positions' latent vectors are replaced with a shared learnable feature vector, we simply replace them with random vectors.
The masked feature encoder output is fed into the contrastive module for producing context vectors.
In parallel, the feature encoder output is also passed to the quantizer without masking to yield its quantized vectors.
For a context vector~$c_{t}$ corresponding to a masked time step~$t$, the model is asked to identify its true quantized vector~$q_{t}$ from a set of~$K$ distractors~$\{\tilde{q}_{1}, \tilde{q}_{2}, ..., \tilde{q}_{K}\}$ that are also quantized vectors uniformly sampled from other masked time steps of the same utterance.
We denote the loss as~$\mathcal{L}_{w}$, and further augment it with a codebook diversity loss~$\mathcal{L}_{d}$ to encourage a uniform usage of codes.
Therefore, the final contrastive loss is defined as:
\begin{equation}
  \mathcal{L}_{c} = \mathcal{L}_{w} + \alpha \cdot \mathcal{L}_{d},
  \label{eq:contrastive_loss}
\end{equation}
where~$\alpha = 0.1$ following~\cite{baevski2020wav2vec}.

\myparagraph{Masked prediction loss}
The context vectors produced by the contrastive module are directly passed to the masked prediction module for producing the final context vectors that are to be used to complete a masked prediction task.
A softmax layer is appended on top of the module's last conformer block.
If a context vector at the final layer corresponds to a masked position, the softmax layer will take the context vector as input and attempt to predict its corresponding token ID, which is assigned earlier in the contrastive module by the quantizer.
We denote the cross-entropy loss for this masked prediction task as~$\mathcal{L}_{m}$.

w2v-BERT is trained to solve the two self-supervised tasks at the same time.
The final training loss to be minimized is:
\begin{equation}
  \mathcal{L}_{p} = \beta \cdot \mathcal{L}_{c} + \gamma \cdot \mathcal{L}_{m}.
\end{equation}
In our experiments, we simply set both~$\beta$ and~$\gamma$ to~1.

\subsection{Fine-tuning}
During fine-tuning we have access to labeled data.
We apply our pre-trained w2v-BERT to two tasks: LibriSpeech and voice search.

The ASR network is a sequence transducer~\cite{graves2012sequence} that consists of a pre-trained w2v-BERT model and a LSTM~\cite{hochreiter1997long} decoder.
We insert a linear layer with Swish activation~\cite{ramachandran2017searching} and batch normalization~\cite{ioffe2015batch} between the pre-trained w2v-BERT model and the LSTM decoder as the projection block.

\section{Experimental Setup}
\label{sec:setup}
\begin{table*}[htbp]
  \caption{WERs(\%) when using the LibriSpeech~960hr subset as supervised data. We compare models trained without any unlabeled data~(Trained from Scratch), trained using Noisy Student Training~(NST) without any pre-training~(Self-training Only), fine-tuned from a pre-trained model only using supervised data~(Pre-training Only), and the models obtained by combining pre-training and self-training~(Pre-training + Self-training). We also include the best results of several methods that we can find from the literature, and their corresponding references are where the numbers are quoted from. The lowest WER(s) under different settings are marked in bold. AM/LM Size denotes the number of parameters in the acoustic/language model. $^{*}$The reason why we do not include Conformer~XL and Conformer~XXL is that, according to~\cite{zhang2020pushing}, simply enlarging Conformer~L produces worse results when the model is trained from scratch. $^{\dagger}$Calculated based on the LM configuration provided in~\cite{xu2021self,baevski2020wav2vec}; $>$ because some information such as the token embedding size is not given therefore not included.}
  \vskip 0.1in
  \label{tab:960hr}
  \centering
  \resizebox{0.9\width}{!}{%
  \begin{tabular}{lccccccccccc}
    \toprule
    \bfseries Method & \multirowcell{2}{\\[-7pt]Unlabeled\\Data~(hrs)}
    & \multirowcell{2}{\\[-7pt]AM\\Size~(B)} & \multirowcell{2}{\\[-7pt]LM\\Size~(B)} & \multicolumn{4}{c}{\bfseries No LM} & \multicolumn{4}{c}{\bfseries With LM} \\
    \cmidrule(r){5-8} \cmidrule(r){9-12}
    & & & & \bfseries dev & \bfseries dev-other & \bfseries test & \bfseries test-other
     & \bfseries dev & \bfseries dev-other & \bfseries test & \bfseries test-other \\
    \midrule
    \bfseries Trained from Scratch \\
    \quad  Conformer~L~\cite{zhang2020pushing}$^{*}$
    & N/A & 0.1 & 0.1
    & 1.9 & 4.4 & 2.1 & 4.3
    & $-$ & $-$ & 1.9 & 3.9 \\
    \midrule
    \bfseries Self-training Only \\
    \quad  Conformer~L with NST~\cite{zhang2020pushing}
    & 60k & 0.1 & 0.1
    & 1.6 & 3.3 & 1.7 & 3.5
    & 1.6 & 3.1 & 1.7 & 3.3 \\
    \midrule
    \bfseries Pre-training Only \\
    \quad wav2vec~2.0~\cite{xu2021self}
    & 60k & 0.3 & $>$ 0.4$^{\dagger}$
    & 2.1 & 4.5 & 2.2 & 4.5
    & 1.6 & 3.0 & 1.8 & 3.3 \\ 
    \quad HuBERT~Large~\cite{hsu2021hubert}
    & 60k & 0.3 & $-$
    & $-$ & $-$ & $-$ & $-$
    & 1.5 & 3.0 & 1.9 & 3.3 \\
    \quad HuBERT~X-Large~\cite{hsu2021hubert}
    & 60k & 1.0 & $-$
    &  $-$ & $-$ & $-$ & $-$
    &  1.5 & \textbf{2.5} & 1.8 & 2.9 \\
    \quad w2v-Conformer~XL~\cite{zhang2020pushing}
    & 60k & 0.6 & 0.1
    & 1.7 & 3.5 & 1.7 & 3.5
    & 1.6 & 3.2 & \bfseries 1.5 & 3.2 \\
    \quad w2v-Conformer~XXL~\cite{zhang2020pushing}
    & 60k & 1.0 & 0.1
    & 1.6 & 3.2 & 1.6 & 3.3 
    & 1.5 & 3.0 & \bfseries 1.5 & 3.1 \\
    \quad w2v-BERT~XL~(Ours)
    & 60k & 0.6 & 0.1
    & \textbf{1.5} & 2.9 & \bfseries 1.5 & 2.9
    & \bfseries 1.4 & 2.8 & \bfseries 1.5 & 2.8 \\
    \quad w2v-BERT~XXL~(Ours) 
    & 60k & 1.0 & 0.1
    & \textbf{1.5} & \textbf{2.7} & \textbf{1.5} & \textbf{2.8}
    & \textbf{1.4} & {2.6} & \textbf{1.5} & \textbf{2.7} \\
    \midrule
    \bfseries Pre-training + Self-training \\
    \quad  wav2vec~2.0~\cite{xu2021self}
    & 60k & 0.3 & $>$ 0.4
    & \bfseries 1.3 & 3.1 &  1.7 & 3.5
    & \bfseries 1.1 & 2.7 & 1.5 & 3.1 \\
    \quad  w2v-Conformer~XXL~\cite{zhang2020pushing}
    & 60k & 1.0 & 0.1
    &  \bfseries 1.3 & 2.7 & 1.5 & 2.8
    &  1.3 & 2.6 & \bfseries 1.4 & 2.7 \\
    \quad  w2v-Conformer~XXL+~\cite{zhang2020pushing}
    & 60k & 1.1 & 0.1
    &  \bfseries 1.3 & 2.7  & 1.5  & 2.7 
    &  1.3 &  2.6 & \bfseries 1.4 &  2.6 \\
    \quad  w2v-BERT~XL~(Ours)
    & 60k & 0.6 & 0.1
    &  \bfseries 1.3 & 2.6 & \bfseries 1.4 & 2.7
    &  1.3 & 2.6 & \bfseries 1.4 & 2.6 \\
    \quad  w2v-BERT~XXL~(Ours)
    & 60k & 1.0 & 0.1
    &  1.4 & \bfseries 2.4  & \bfseries 1.4  & \bfseries 2.5 
    &  1.3 & \bfseries 2.4 & \bfseries 1.4 & \bfseries 2.5 \\
    \bottomrule
  \end{tabular}
  }
\end{table*}

Apart from the pre-training method, the rest of the experimental pipeline follows the exact same setup as in~\cite{zhang2020pushing}.

\subsection{Data}
We use the Libri-Light unlab-60k subset~\cite{kahn2020libri}, which contains about~60,000 hours of unannotated speech audio, for pre-training w2v-BERT models.
For our main results, we use the LibriSpeech 960hr subset~\cite{panayotov2015librispeech} as the supervised data, and use the~100hr subset for ablation studies.
We report word error rates~(WERs) on the dev-clean, dev-other, test-clean, and test-other evaluation subsets.
80-dimensional log-mel filter bank coefficients are used as acoustic inputs to our model.
For transcript tokenization, we use a~1024-token WordPiece model~\cite{schuster2012japanese} that is constructed from the transcripts of the LibriSpeech training set~(or the~100hr subset when the models are fine-tuned on it).

\subsection{Pre-training details}
\myparagraph{Masking} For masking the feature encoder output, we randomly sample the starting positions to be masked with a probability of~0.065 and mask the subsequent~10 time steps~(same as~\cite{baevski2020wav2vec,zhang2020pushing}).
The masked spans may overlap.

\myparagraph{Optimization} We pre-train two versions of w2v-BERT models, one has about~0.6 billion parameters and the other has about~1 billion parameters, denoted as w2v-BERT~XL and w2v-BERT~XXL, respectively.
The two variants share the same model configuration that is summarized in Table~\ref{tab:model_params}, and their only difference is the number of conformer blocks.
Specifically, w2v-BERT~XL's contrastive module consists of~12 conformer blocks and the masked prediction module is composed of another~12.
w2v-BERT~XXL, while having the same amount of conformer blocks in its contrastive module, enlarges its masked prediction module to~30 conformer blocks.
For w2v-BERT~XL, we train it with a batch size of~2048 using the Adam optimizer~\cite{kingma2015adam} with a transformer learning rate schedule as described in section~5.3 of~\cite{vaswani2017attention}.
The peak learning rate is~2e-3 and the warm-up steps are~25k.
For w2v-BERT-XXL, we train it with the Adafactor optimizer~\cite{shazeer2018adafactor} with~$\beta_{1} = 0.9$ and~$\beta_{2} = 0.98$, with the learning rate schedule remaining the same.

\subsection{Fine-tuning details}
\myparagraph{Optimization} For both w2v-BERT~XL and w2v-BERT-XXL, we take their pre-trained checkpoints at~400k steps, and fine-tune them on the supervised data with a batch size of 256.
The decoder for both models are a two-layer LSTM with a hidden dimension of~640.
We employ separate optimizers and learning rate schedules for optimizing the pre-trained model and the decoder, given the fact that the former has been pre-trained while the latter needs to be trained from scratch.
For w2v-BERT~XL, both the pre-trained model and the decoder are optimized with an Adam optimizer with a transformer learning schedule.
The difference is that for the pre-trained component we use a peak learning rate of~3e-4 with~5k warm-up steps, while for the decoder we use a peak learning rate of~1e-3 and~1.5k warm-up steps.
For w2v-BERT-XXL, an Adafactor optimizer that has the same configuration as in pre-training is used, and the learning rate schedules for the encoder and decoder are the same as the XL variant.

\myparagraph{Self-training, data augmentation, and LM fusion}
In addition to self-supervised pre-training, in the fine-tuning stage we also employ a number of practical techniques that further improve models' performance on ASR.
These techniques include SpecAugment~\cite{park2019specaugment,park2020specaugment} for data augmentation, Noisy Student Training~\cite{park2020improved} for self-training, and language model fusion for decoding.
When any of the techniques are used, we follow the exact same setup as in~\cite{zhang2020pushing}.
We refer the readers to the paper for the details on these techniques.

\section{Results and Discussion}
\label{sec:results}

\subsection{Main results}
In Table~\ref{tab:960hr}, we present our results on the four LibriSpeech evaluation sets using the~960hr subset as the supervised data.
We compare w2v-BERT to a number of state-of-the-art self-supervised representation learning methods from the literature such as HuBERT~\cite{hsu2021hubert} and wav2vec~2.0~\cite{baevski2020wav2vec} under different semi-supervised settings, including whether self-training is employed during the fine-tuning stage and whether a language model is incorporated during inference time.
We also include the model size of the ASR network used by each method, denoted as acoustic model~(AM) Size and language model~(LM) Size.
Results missing from the literature~(e.g., results of HuBERT without self-training and LM) are indicated with a~``$-$'' in the table.
From Table~\ref{tab:960hr} we have the following two key conclusions.

\myparagraph{{\it \textbf{Without self-training and LM, w2v-BERT already either outperforms or matches other models with LM.}}}
We see that with just pre-training when neither self-training nor LM is used, w2v-BERT~XL achieves a WER of~1.5/2.9~(test/test-other), which already either outperforms or matches other models with LM, and outperforms their counterparts without LM by a larger margin.
w2v-BERT~XXL further increases the gap on the more challenging dev-other and test-other subsets.
Noticeably, compared to wav2vec~2.0, w2v-BERT-XXL shows a relative WER reduction of~28\%,~42\%,~32\%, and~38\% on the four evaluation subsets respectively without LM, and~13\%/~13\%/~17\%/~18\% when LM is employed.

We want to highlight that although w2v-BERT~XL~(0.6B) and w2v-BERT~XXL~(1.0B) have a larger pre-trained model size than wav2vec~2.0~(0.3B), the latter also incorporates a much larger LM~($>$ 0.4B) during self-training and decoding according to~\cite{xu2021self,baevski2020wav2vec}.
When considering the sum of the two components, wav2vec~2.0~($>$ 0.7B) actually features a similar~(if not bigger) model size as w2v-BERT~XL~(0.7B).

\begin{figure*}[htbp]
  \centering
  \begin{subfigure}[b]{0.325\textwidth}
    \centering
    \includegraphics[width=\columnwidth]{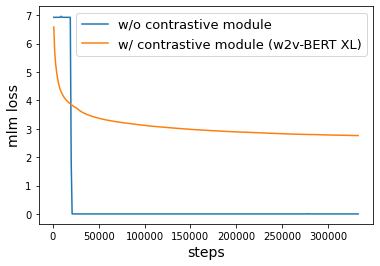}
    \caption{MLM training loss}
    \label{fig:mlm_loss}
  \end{subfigure}
  \hfill
  \begin{subfigure}[b]{0.33\textwidth}
    \centering
    \includegraphics[width=\columnwidth]{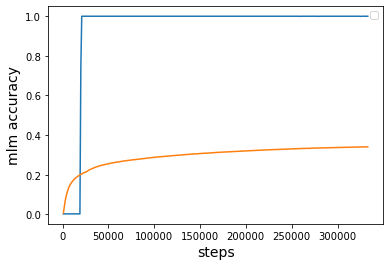}
    \caption{MLM training accuracy}
    \label{fig:mlm_acc}
  \end{subfigure}
  \hfill
  \begin{subfigure}[b]{0.33\textwidth}
    \centering
    \includegraphics[width=\textwidth]{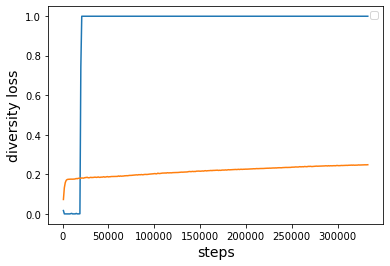}
    \caption{Training diversity loss}
    \label{fig:div_loss}
  \end{subfigure}
  \caption{Training curves of w2v-BERT models with and without contrastive module. From left to right: MLM training loss, MLM training accuracy, training diversity loss. The blue curve represents the w2v-BERT model without contrastive module, and the orange curve represents w2v-BERT~XL~(with contrastive module). We show results for the first 300k steps.}
  \label{fig:learning_curves}
\end{figure*}

\myparagraph{{\it \textbf{Contrastive learning combined with masked language modeling is more effective than contrastive learning alone.}}}
w2v-Conformer~\cite{zhang2020pushing} and w2v-BERT only differ in their pre-training method and have all other aspects in common such as their model size and fine-tuning pipeline.
This allows a truly apple-to-apple comparison between the two pre-training methods for their effectiveness in representation learning.
Below we briefly describe their differences in pre-training.

w2v-Conformer adopted wav2vec~2.0's contrastive task as the sole pre-training objective, but replaced the quantization module with a linear layer as the authors did not find quantization helpful for improving downstream ASR performance.
w2v-BERT, on the other hand, adopts wav2vec~2.0's contrastive task not just for learning contextualized speech representations, but mainly for the purpose of obtaining a codebook that can represent every segment of continuous speech as an discriminative discrete token, such that we can exploit MLM for learning powerful speech representations.
As will be demonstrated in our analysis~(Section~\ref{sec:analysis}), the contrastive loss is essential for making MLM to work.

By comparing the Pre-training Only results of w2v-Conformer and w2v-BERT, we see that w2v-BERT~XL, despite having fewer model parameters, already outperforms w2v-Conformer-XXL---the previous state of the art---especially on the dev-other and test-other subsets.
When self-training is applied, w2v-BERT~XL still either outperforms or matches w2v-Conformer-XXL's results.
w2v-BERT-XXL, which is of the same model size as w2v-Conformer~XXL, outperforms w2v-Conformer~XXL on even more evaluation subsets.
These results demonstrate the superiority of the proposed w2v-BERT over existing pre-training frameworks.


\subsection{Analysis and discussion}
\label{sec:analysis}
The goal of our analysis is to understand the roles of contrastive learning as well as its learned codebook
in the w2v-BERT pre-training framework.

\myparagraph{On the necessity of contrastive module}
The first natural question is whether the contrastive module is an essential component of the framework, or is that a masked prediction module alone can already derive a suitable codebook for its own MLM purpose.

Without the contrastive module, the feature encoder output is directly fed to the masked prediction module.
Intuitively, the masked prediction module then gets a full control over the quantizer~(which may originally be viewed as part of the contrastive module) and decide its own prediction target.
In order to maximize the prediction performance, the masked prediction module can ``cheat'' by coming up with a trivial solution where it asks the quantizer to cooperate with it by quantizing all feature encoder's output frames that correspond to the masked positions to the same code vector, in other words, always assigning the same target ID for the masked prediction module to predict.
The module thus perfectly solves the masked prediction task without learning any useful representation.

To verify our intuition, we train a series of w2v-BERT models without the contrastive module.
These variants all have the same capacity as w2v-BERT~XL, that is, they are all constructed with~24 conformer layers.
We try different values of~$\alpha = 0.1, 0.3, 0.5$, and~$0.7$ in Equation~\ref{eq:contrastive_loss} for increasing encouragement of uniform usage of codes.
Nevertheless, we find all models are untrainable and quickly collapse regardless of the value of~$\alpha$.
In Figure~\ref{fig:learning_curves} we show the training curves of a w2v-BERT model without contrastive module with~$\alpha$ set to~0.5.
We include the curves of the successfully trained w2v-BERT~XL for comparison.
The plots include the models' masked prediction loss~(Figure~\ref{fig:mlm_loss}), masked prediction accuracy~(Figure~\ref{fig:mlm_acc}), and diversity loss~(Figure~\ref{fig:div_loss}) during pre-training.

We find that the training curves of w2v-BERT without contrastive module~(in blue) strongly align with our intuition: the masked prediction loss quickly decreases to close to~0 at the early stage of training~(Figure~\ref{fig:mlm_loss}) where the model reaches~100\% prediction accuracy~(Figure~\ref{fig:mlm_acc}).
Meanwhile, as shown in Figure~\ref{fig:div_loss}, the diversity loss quickly increases to close to~1, where in our implementation this indicates an extremely low entropy of softmax distribution over the codebook entries, suggesting code collapse.
Comparing the curves of w2v-BERT models with and without contrastive module, we hypothesize that the contrastive loss guides the entries in the codebook to be discriminative, thus preventing the masked prediction module from deriving a trivial solution just to maximize masked prediction performance.

\myparagraph{On the impact of contrastive module's capacity}
After confirming the necessity of contrastive module, next we are interested in investigating the impact of its capacity on downstream ASR performance.

We train a series w2v-BERT models with different numbers of conformer layers in their contrastive module.
To rule out the factor of masked prediction module's capacity, we keep the total number of conformer layers in the two modules fixed at~24.
We use~$C_{n}$ to denote each variant, where~$n$ is the number of conformer layers in the contrastive module.
For instance,~$C_{4}$ has~4 conformer layers in its contrastive module and~$20$ in its masked prediction module.
Here we consider~$n = 2, 4, 6, 8, 10, 12,$ and~$24$, where~$C_{24}$ is an extreme case where the two modules are completely overlapped with each other and hence the contrastive and MLM tasks will be both tackled at the last~(24-th) layer.
Note that~$C_{12}$ is essentially w2v-BERT~XL.

We use the LibriSpeech~100hr subset as the supervised data for this experiment, and both self-training and LM fusion are {\it not} used when training the ASR network.
Results are shown in Table~\ref{tab:100hr}.
We include the results of some pre-training methods from the literature that also do not incorporate self-training and LM.

\begin{table}[htbp]
  \caption{WERs~(\%) when using the LibriSpeech~100hr subset as supervised data. For all methods, both self-training and LM fusion are not used. References are where the numbers are quoted from.}
  \vskip 0.1in
  \label{tab:100hr}
  \centering
  \resizebox{0.95\width}{!}{%
  \begin{tabular}{lcccc}
    \toprule
    \bfseries Method & \bfseries dev & \bfseries dev-other & \bfseries test & \bfseries test-other \\
    \midrule
    \bfseries Baseline \\
    \quad  wav2vec~2.0~\cite{baevski2020wav2vec}
    & 3.3 & 6.5 & 3.1 & 6.3\\
    \quad  w2v-Conformer~XL~\cite{zhang2020pushing}
    & 2.5 & 4.7 & 2.6 & 4.9\\
    \quad  w2v-BERT~XXL~(Ours)
    & 2.3 & 4.0 & 2.3 & 4.3 \\
    \midrule
    \bfseries w2v-BERT w/~24 layers \\
    \quad  $C_{2}$
    & 2.4 & 5.1 & 2.5 & 5.1 \\
    \quad  $C_{4}$
    & 2.5 & 4.6 & 2.5 & 5.1 \\
    \quad  $C_{6}$
    & 2.5 & 4.2 & 2.4 & 4.7 \\
    \quad  $C_{8}$
    & 2.3 & 4.3 & 2.4 &  4.6 \\
    \quad  $C_{10}$
    & 2.4 & 4.5 & 2.5 & 4.8 \\
    \quad  $C_{12}$~(w2v-BERT~XL)
    & 2.4 & 4.4 & 2.5 & 4.6 \\
    \quad  $C_{24}$
    & 2.4 & 4.9 & 2.5 & 5.0 \\
    \bottomrule
  \end{tabular}
  }
\end{table}

From Table~\ref{tab:100hr} we can roughly observe a performance sweet spot on all four evaluation subsets when we increase the number of layers in the contrastive module.
From~$C_{2}$ to~$C_{8}$, the WERs are mostly decreasing, meaning that enlarging the contrastive module is helpful for learning better representations.
The fact that the performance continues to improve while the masked prediction module shrinks~(and hence becomes less expressive) as we deepen the contrastive module further suggests the importance of making the contrastive module sufficiently large.

Starting from~$C_{8}$, however, the WERs stop decreasing as we deepen the contrastive module.
We hypothesize that this is because the masked prediction module has now become too small to learn representations useful for the MLM task.
Such reasoning is supported by the fact that enlarging the masked prediction module while keeping the contrastive module the same size can still improve the performance~(w2v-BERT~XL vs. w2v-BERT~XXL).

Last but not least, we see that w2v-BERT always outperforms wav2vec~2.0 regardless of its layer configuration.
It also either outperforms or matches w2v-Conformer~XL's performance when its contrastive module has enough capacity~(i.e., when~$n > 4$).



\subsection{Results on Voice Search traffic}



So far we have shown w2v-BERT pre-trained on read speech audio can achieve great performance on the well-benchmarked LibriSpeech task.
To validate the effectiveness of w2v-BERT on real-world audio traffic, we apply it to Google's Voice Search traffic.
Our train and test sets are derived from~\cite{li2021scaling}.
We use~34.3k hours of English audio for pre-training, and randomly pick~1k hours as the fine-tuning data, which is anonymized and human-transcribed.
The test set contains around~12k Voice Search utterances with duration less than~5.5s long.
The testing utterances are anonymized and human-transcribed, and are representative of Google’s Voice Search traffic.

The traffic data is more challenging to be used for pre-training than read speech audio in two folds: (1) It is noisier and contains more silences that make negative sampling for contrastive learning less effective.
(2) The average length of the traffic audio~(5 seconds) is much shorter than that of read speech audio.
These factors make the context learned from the audio segments much less effective.

\begin{table}[htbp]
  \caption{Results on voice search data. Baseline conformer model is 100M parameters. All the other models are 600M parameters, marked as XL.}
  \vskip 0.1in
  \label{tab:vs}
  \centering
  \resizebox{0.9\width}{!}{%
  \begin{tabular}{lccc}
    \toprule
    \bfseries Method & Unlabeled Data~(Domain) & \bfseries Test~(VS) \\
    \midrule 
    \quad  Conformer  & N/A  & 10.7 \\
    \midrule
    \quad w2v-Conformer-XL & 34.3k~(Voice search) & 10.8 \\
    \quad w2v-Conformer-XL-tuned & 34.3k~(Voice search) & 8.9 \\
    \quad  w2v-BERT XL~(Ours) & 34.3k~(Voice search) & \textbf{6.2} \\
    \bottomrule
  \end{tabular}
  }
\end{table}

As shown in Table~\ref{tab:vs}, if we take the same training script as w2v-Conformer-XL, the model tends to cheat on negative samples due to the large portion of non-speech and shorter context.
To make contrastive learning more effective, we have to use a less aggressive subsampling: instead of using a~4 times convolutional stride, we stack~3 frames as target to encourage the model to learn better context.
However, by taking an identical architecture and using the same training receipt, our w2v-BERT~XL significantly improves the tuned contrastive baseline by relative~30\%.

\section{Conclusion and Future Work}
\label{sec:conclusion}
We proposed w2v-BERT for self-supervised speech representation learning.
w2v-BERT is composed of a contrastive module for discretizing continuous speech and a masked prediction module that performs masked language modeling with the discretized speech.
The two modules can be jointly optimized.
We pre-train w2v-BERT on~60k hours of unlabeled speech data from the Libri-Light corpus, and show it either outperforms or matches state-of-the-art systems such as w2v-Conformer, HuBERT, and wav2vec~2.0.
The gain also transfers to a more challenging internal dataset.
We also provide an analysis on the importance of the contrastive module for enabling effective masked language modeling in w2v-BERT.

In our experiments, all the hyperparameter setups are directly taken from~\cite{zhang2020pushing} without any changes.
For future work, we plan to first search for the best training configuration for w2v-BERT.
We are also interested in evaluating w2v-BERT in low-resource data settings using the Libri-Light~10min,~1hr, and~10hr benchmarks.


\bibliographystyle{IEEEbib}
\bibliography{refs,strings}

\end{document}